\documentclass[final]{cvpr}

\usepackage{times}
\usepackage{epsfig}
\usepackage{graphicx}
\usepackage{amsmath}
\usepackage{amssymb}

\usepackage[pagebackref=true,breaklinks=true,colorlinks,citecolor=darkblue,bookmarks=false]{hyperref}

\usepackage{enumitem,amsmath,amssymb,amsthm,commath,dsfont,complexity,mathtools,nicefrac,booktabs,graphicx}

\def\paragraph#1{\medskip\noindent\textbf{\boldmath #1\,}}

\usepackage{xcolor}
\definecolor{darkblue}{rgb}{0,0.08,0.45}
\definecolor{seaborn_blue}{HTML}{0072B2}
\definecolor{seaborn_green}{HTML}{009E73}
\def\gain#1{\textbf{\textcolor{seaborn_blue}{#1}}}

\def\cvprPaperID{6392} 
\def\confYear{CVPR 2021}

\begin{document}

\title{DAP: Detection-Aware Pre-training with Weak Supervision}

\newcounter{@inst}
\def\inst#1{\unskip$^{#1}$}
\def\email#1{{\small\tt#1}}

\author{
Yuanyi Zhong\inst{1},
Jianfeng Wang\inst{2},
Lijuan Wang\inst{2},
Jian Peng\inst{1},
Yu-Xiong Wang\inst{1},
Lei Zhang\inst{2}
\\ \\
\inst{1} University of Illinois at Urbana-Champaign \email{\{yuanyiz2, jianpeng, yxw\}@illinois.edu}
\\
\inst{2} Microsoft \email{\{jianfw, lijuanw, leizhang\}@microsoft.com}
}

\maketitle

\begin{abstract}
This paper presents a detection-aware pre-training (DAP) approach, which leverages only weakly-labeled classification-style datasets (e.g., ImageNet) for pre-training, but is specifically tailored to benefit object detection tasks. In contrast to the widely used image classification-based pre-training (e.g., on ImageNet), which does not include any location-related training tasks, we transform a classification dataset into a detection dataset through a weakly supervised object localization method based on Class Activation Maps to directly pre-train a detector, making the pre-trained model location-aware and capable of predicting bounding boxes. We show that DAP can outperform the traditional classification pre-training in terms of both sample efficiency and convergence speed in downstream detection tasks including VOC and COCO. In particular, DAP boosts the detection accuracy by a large margin when the number of examples in the downstream task is small.
\end{abstract}

\section{Introduction}

\begin{figure}[t]
    \centering
    \includegraphics[width=\linewidth]{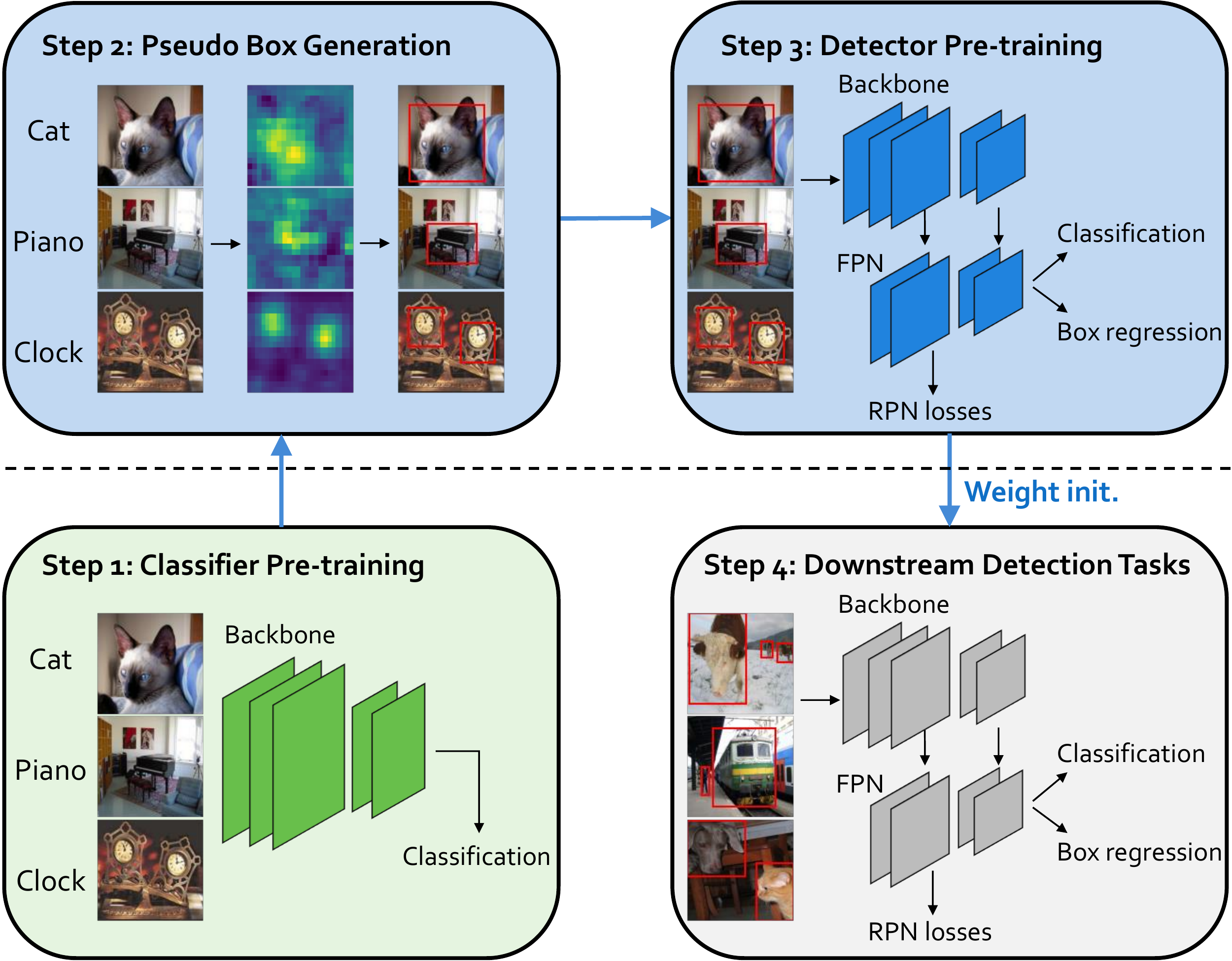}
    \caption{The DAP workflow. It consists of 4 steps: (1) Classifier pre-training on a weak supervision dataset, (2) Pseudo box generation by WSOL (\eg, through CAM as illustrated), (3) Detector pre-training with the generated pseudo boxes, (4) Downstream detection tasks. The traditional classification pre-training and fine-tuning directly go from Step (1) to (4) at the bottom, while DAP inserts the additional Steps (2) and (3) at the top. In both cases, the pre-trained weights are used to initialize the downstream models. DAP gives the model a chance to learn how to perform explicit localization, and is able to pre-train detection-related components while classification pre-training cannot, such as the FPN, RPN, and box regressor in a Faster RCNN detector.}
    \label{fig:workflow}
    \vspace{-10pt}
\end{figure}

Pre-training and fine-tuning have been a dominant paradigm for deep learning-based object recognition in computer vision \cite{girshick2014rich,donahue2014decaf,ren2015faster,he2017mask}.
In such a paradigm, neural network weights are typically pre-trained on a large dataset (\eg, through ImageNet \cite{deng2009imagenet} classification training), and then transferred to initialize models in downstream tasks. Pre-training can presumably help improve downstream tasks in multiple ways. The low-level convolutional filters, such as edge, shape, and texture filters, are already well-learned in pre-training \cite{zeiler2014visualizing}. The pre-trained network is also capable of providing meaningful semantic representations. For example, in the case of ImageNet classification pre-training, since the number of categories is large (1000 classes), the downstream object categories might be related to a subset of the pre-training categories and can reuse the pre-trained feature representations. Pre-training may also help the optimizer avoid bad local minima by providing a better initialization point than a completely random initialization \cite{erhan2009difficulty}. Therefore, fine-tuning would only require a relatively small number of gradient steps to achieve competitive accuracy.

However, the empirical gain for object detection brought by classification pre-training is diminishing with successively larger pre-training datasets, ranging from ImageNet-1M, ImageNet-5k \cite{he2017mask}, to ImageNet-21k (14M), JFT-300M \cite{sun2017revisiting}, and billion-scale Instagram images \cite{mahajan2018exploring}. 
Meanwhile, \cite{he2019rethinking} shows that training from random initialization (\ie, from scratch) can work equally well with sufficiently large data (COCO~\cite{lin2014microsoft}) and a sufficiently long training time, making the effect of classification pre-training questionable.

We conjecture that the diminishing gain of classification pre-training for object detection is due to several mismatches between the pre-training and the fine-tuning tasks. Firstly, the task objectives of classification and detection are different. Existing classification pre-training is typically unaware of downstream detection tasks. The pre-training adopts a single whole-image classification loss which encourages translation and scale-invariant features, while the detection fine-tuning involves several different classification and regression losses which are sensitive to object locations and scales. Secondly, the data distributions are misaligned. The localization information required by detection is not explicitly made available in classification pre-training. Thirdly, the architectures are misaligned. The network used in pre-training is a bare backbone network such as a ResNet model \cite{he2016deep} followed by an average pooling and a linear classification layer. In contrast, the network in an object detector contains various additional architectural components such as the Region Proposal Network (RPN) \cite{ren2015faster}, the Feature Pyramid Network (FPN) \cite{lin2017feature}, the ROI classification heads and the bounding box regression heads \cite{ren2015faster}, \etc. These unique architectural components in detectors are not pre-trained and are instead randomly initialized in detection fine-tuning, which could be sub-optimal.

Aiming at bridging the gap between pre-training with classification data and detection fine-tuning, we introduce a Detection-Aware Pre-training (DAP) procedure as shown in Figure~\ref{fig:workflow}.
There are two desired properties that are necessary to pre-train a detector: (1) Classification should be done \emph{locally} rather than globally; (2) Features should be \emph{capable} of predicting bounding boxes and can be easily adapted to any desired object categories after fine-tuning.
With the desired properties in mind, DAP starts from pre-training a classifier on the classification data, and extracts the localization information with existing tools developed in Weakly Supervised Object Localization (WSOL) based on Class Activation Maps (CAM)~\cite{zhou2016learning}. 
The next step is to treat the localized instances as pseudo bounding boxes to pre-train a detection model. Finally, the pre-trained weights are used for model initialization in downstream detection tasks such as VOC \cite{everingham2010pascal} and COCO \cite{lin2014microsoft}. 
DAP enables the pre-training of (almost) the entire detector architecture and offers the model the opportunity to adapt its representation to perform localization explicitly.
Our problem setting focuses on leveraging the weak image-level supervision in classification-style data for pre-training (ImageNet-1M and ImageNet-14M) \cite{deng2009imagenet}, therefore makes a head-to-head comparison to the traditional classification pre-training.
Note that our setting is different from unsupervised pre-training \cite{he2020momentum,chen2020simple,chen2020big} which is only based on unlabeled images, and is different from fully-supervised detection pre-training \cite{shao2019objects365} which is hard to scale.

Comprehensive experiments demonstrate that adding the simple lightweight DAP steps in-between the traditional classification pre-training and fine-tuning stages yields consistent gains across different downstream detection tasks. The improvement is especially significant in the low-data regime. This is particularly useful in practice to save the annotation effort. In the full-data setting, DAP leads to faster convergence than classification pre-training and also improves the final detection accuracy by a decent margin. Our work suggests that a carefully designed detection-specific pre-training strategy with classification-style data can still benefit object detection. We believe that this work makes the first attempt towards detection-aware pre-training with weak supervision.
\section{Related Work}

\vspace{-6pt}
\paragraph{Pre-training and fine-tuning paradigm.}
Pre-training contributed to many breakthroughs in applying CNN for object recognition \cite{girshick2014rich,donahue2014decaf,ren2015faster,he2017mask}. A common strategy, for example, is to pre-train the networks through supervised learning on the ImageNet classification dataset \cite{deng2009imagenet,russakovsky2015imagenet} and then fine-tune the weights in downstream tasks. 
Zeiler \etal visualize the convolutional filters in a pre-trained network, and find that intermediate layers can capture universal local patterns, such as edges and corners that can be generalizable to other vision tasks \cite{zeiler2014visualizing}. Pre-training may ease up the difficult optimization problem of fitting deep neural nets via first-order methods \cite{erhan2009difficulty}.
Recently, the limit of supervised pre-training has been pushed by scaling up the datasets. In Big Transfer (BiT), the authors show that surprisingly high transfer performance can be achieved across 20 downstream tasks by classification pre-training on a dataset of 300M noisy-labeled images (JFT-300M) \cite{chen2020big}. Notably, pre-training on JFT-300M drastically improves the performance with small data.
Similarly, Mahajan \etal explore the limits of (weakly) supervised pre-training with noisy hashtags on billions of social media (Instagram) images \cite{mahajan2018exploring}. The traditional ImageNet-1M becomes a small dataset compared to the Instagram data. A gain of $5.6\%$ can be achieved on ImageNet-1M classification accuracy by pre-training on the billion-scale data.
As for related work in other deep learning fields, pre-training is also a dominant strategy in natural language processing (NLP) and speech processing \cite{schneider2019wav2vec,xu2020self}. For example, BERT \cite{devlin2019bert} and GPT-3 \cite{brown2020language} show that language models pre-trained on massive corpora can generalize well to various NLP tasks.

\paragraph{Pre-training and object detection.}
However, the story of how and to what extent classification pre-training is helping object detection is up for debate. On one hand, it is observed that pre-training is important when downstream data is limited \cite{agrawal2014analyzing,he2019rethinking}. On the other hand, there is a line of work reporting competitive accuracy when training modern object detectors from scratch \cite{szegedy2013deep,shen2017dsod,zhu2019scratchdet,he2019rethinking}. The gain brought by classification pre-training on larger datasets seems diminishing \cite{kolesnikov2019big,mahajan2018exploring,he2019rethinking}. Classification pre-training may sometimes even harm localization when the downstream data is abundant while benefit classification \cite{mahajan2018exploring}.
Shinya \etal try to understand the impact of ImageNet classification pre-training on detection and discover that the pre-trained model generates narrower eigenspectrum than the from-scratch model \cite{shinya2019understanding}.
Recent work proposes a cheaper Montage pre-training for detection on the \emph{target detection} data and obtains an on-par or better performance than ImageNet classification pre-training \cite{zhou2020cheaper}. Our work aims at improving the usefulness of pre-training with \emph{classification-style} data (\eg, ImageNet) for detection, by resolving the misalignment between pre-training and fine-tuning tasks through the Detection-Aware Pre-training procedure. 
Leveraging weak supervision is encouraging as the pre-training dataset can be easily scaled up.
This is different from pre-training on a fully-supervised detection data \cite{shao2019objects365,li2019analysis}, which requires expensive annotation cost.

\paragraph{Weakly Supervised Object Localization (WSOL).}
We leverage WSOL in DAP to locate bounding boxes. WSOL refers to a class of object localization methods that rely on weak supervision (image-level labels) \cite{oquab2015object,zhou2016learning,singh2017hide,zhang2018adversarial,zhang2018self,choe2019attention,choe2020evaluating}, which is exactly what we need for the pre-training data. Many of those methods are based on Class Activation Maps (CAMs) \cite{oquab2015object,zhou2016learning,zhang2018self}.
CAMs highlight the strongest activation regions for a given class thus can roughly locate objects. CAM-style methods remain among the most competitive approaches for WSOL to date \cite{choe2020evaluating}. Weakly Supervised Object Detection (WSOD) \cite{bilen2016weakly,kantorov2016contextlocnet,tang2017multiple,wei2018ts2c,zeng2019wsod2,zhong2020boosting} is a highly related area to WSOL. WSOD tends to focus on detecting possibly multiple objects in multi-labeled images, while WSOL focuses on localizing one object instance. 
Comparably, WSOD requires more computational cost, and thus we focus on WSOL for large-scale pre-training data.

\paragraph{Self-supervised learning.} 
Self-supervised (\eg, the contrastive learning approaches \cite{he2020momentum,chen2020simple,chen2020big,erhan2010does}) pre-training utilizes raw images to pre-train a network without any annotation.
While this is an emerging area, the task is challenging due to the lack of annotations, especially for object detection. 
For example, the backbone in these works still shares the same backbone with the classification task, and ignores detection-related components, \eg, feature pyramid network.
Meanwhile, the goal of these works is different from ours. 
We target at leveraging \emph{classification}-style data \emph{specifically} for detection, while they focus on learning general visual representation from unlabeled data.

\paragraph{Self-training.}
Self-training, which refers to the technique of iterative pseudo-labeling and re-training in semi-supervised learning, can also improve detection performance \cite{radosavovic2018data,zoph2020rethinking}.
Self-training \cite{zoph2020rethinking} revisits a large auxiliary dataset multiple times, while we assume that the pre-training dataset is not available in downstream tasks. 
In addition, self-training is complementary to an improved pre-training approach, which has been verified in speech recognition \cite{xu2020self} and will also be demonstrated in Sec.~\ref{sec:exp}.

\section{Detection-Aware Pre-training}

\subsection{Overview of workflow}

Figure~\ref{fig:workflow} illustrates the workflow of Detection-Aware Pre-training (DAP) with image-level annotations for object detection. We describe each step in detail below.

\paragraph{Step 1: Classifier Pre-training.}
The foremost step is to train a deep CNN classifier on the pre-training dataset. Deep CNN classifier usually connects a CNN backbone with an average pooling layer and a linear classification layer \cite{he2016deep}. The network is typically trained with a cross-entropy loss on the image and image-level label pairs \cite{he2016deep}.

The traditional classification pre-training approach directly transfers the network weights of the backbone into the downstream detection fine-tuning tasks. DAP adds the pseudo box generation and the detector pre-training steps in between. In both pre-training approaches, the neural network weights are the only medium of knowledge transfer.

\begin{figure}[t]
    \centering
    \includegraphics[width=\linewidth]{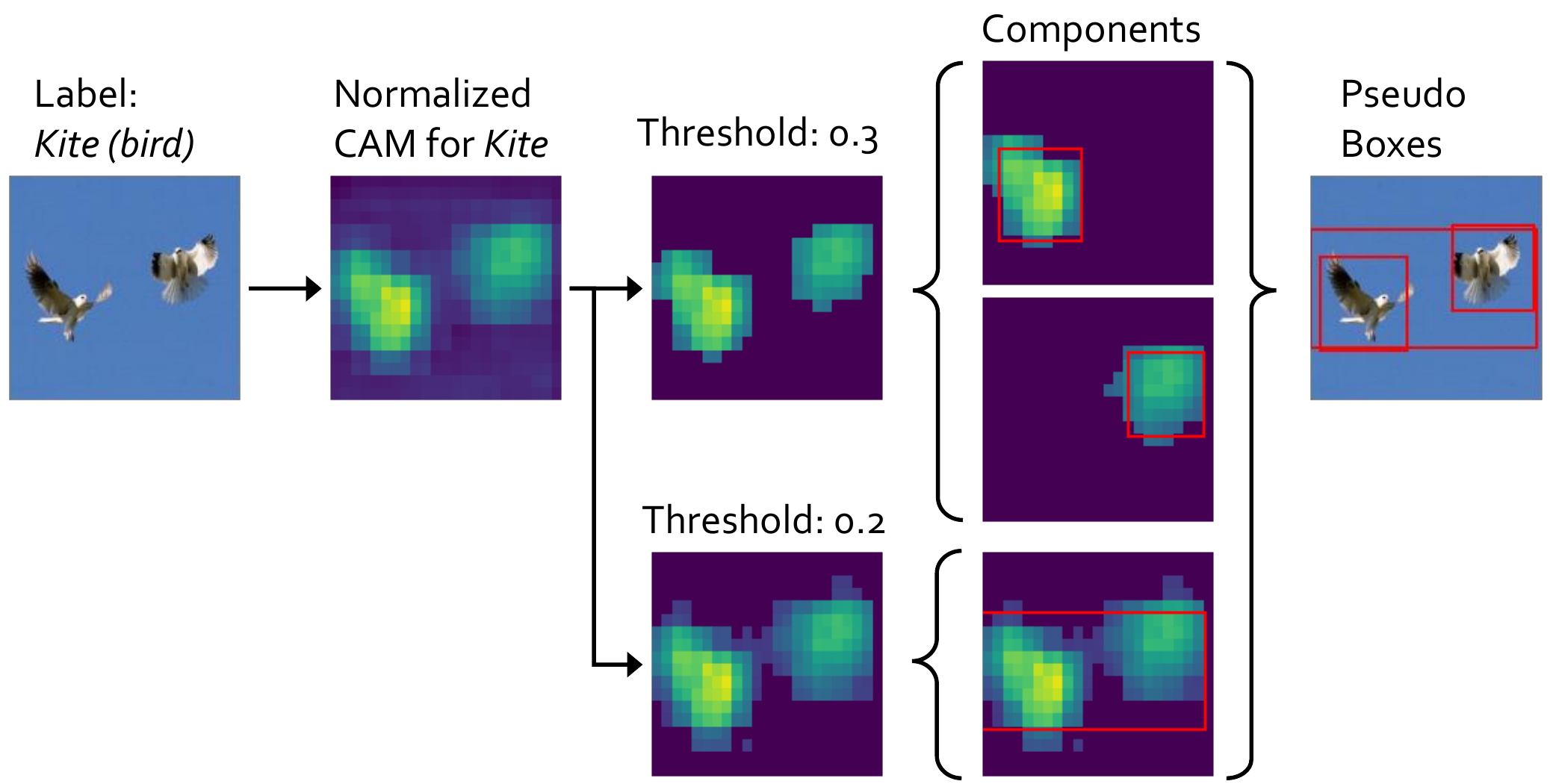}
    \caption{Pseudo box generation procedure. We threshold the CAM with different values and fit a box for each large connected component in each thresholded CAM. The results are merged with NMS. The reason to use different thresholds is to increase the recall. In this example, a too low threshold would fail to discern the two kites. In other cases, a too high threshold might fail to capture the object's whole extent. We find these noisy pseudo labels are sufficient to pre-train the detector to achieve noticeable gains.}
    \label{fig:cam_method}
    \vspace{-10pt}
\end{figure}

\paragraph{Step 2: Pseudo Box Generation with CAM.}
From a trained CNN classifier, the Class Activation Map (CAM) of a ground-truth labeled class can be extracted by converting the final classification layer of that class directly into a $1\times 1$ convolution on the last feature map with the average pooling layer removed (and without the activation function) \cite{oquab2015object,zhou2016learning}. To improve quality, we can average the CAMs obtained from
images with different transformations, \eg, left-right flip and multi scales.

We develop a simple procedure inspired by existing WSOL literature \cite{oquab2015object,zhou2016learning,zhang2018self,choe2020evaluating} to infer bounding boxes from a CAM, as illustrated in Figure~\ref{fig:cam_method}. 
First, the CAM is normalized to range $[0,1]$ via an affine transformation based on the extreme values. Here $x, y$ are the horizontal and vertical coordinates:
\begin{equation}
    \mathrm{CAM}(x,y) = \frac
      {\mathrm{CAM}(x,y) - \min \mathrm{CAM}(x,y)}
      {\max \mathrm{CAM}(x,y) - \min \mathrm{CAM}(x,y)}.
    \label{eq:cam_norm}
\end{equation}
Then we threshold the CAM with a hyper-parameter $\tau$ and an indicator function $\mathds{1}\{\cdot\}$:
\begin{equation}
    M(x,y) = \mathrm{CAM}(x,y) \times \mathds{1}\{ \mathrm{CAM}(x,y) > \tau \}.
    \label{eq:cam_thresh}
\end{equation}

Several object instances of the same category could present in a single image, \eg, the two kites in Figure~\ref{fig:cam_method}. Hence we find connected components on the thresholded CAM $M$ and filter out the components if the area is less than half of the largest component's area. This could remove noisy and small components.
Then, we calculate the bounding box coordinates for each component. 
Denote $\Omega$ as the point set of one component. The bounding box $(x_c, y_c, w, h)$ covering $\Omega$ is constructed by matching the first and second moments (mean and variance) with a rectangle through the following equations:
\begin{align}
    x_c &= \frac{ \sum_{(x,y) \in \Omega} M(x,y) x }{ \sum_{(x,y) \in \Omega} M(x,y) }
    ,
    \\
    y_c &= \frac{ \sum_{(x,y) \in \Omega} M(x,y) y }{ \sum_{(x,y) \in \Omega} M(x,y) }
    ,
    \\
    w &= \sqrt{
    12 \frac
       { \sum_{(x,y) \in \Omega} M(x,y) (x - x_c)^2 }
       { \sum_{(x,y) \in \Omega} M(x,y) }
    }
    ,
    \\
    h &= \sqrt{
    12 \frac
       { \sum_{(x,y) \in \Omega} M(x,y) (y - y_c)^2 }
       { \sum_{(x,y) \in \Omega} M(x,y) }
    }
    .
\end{align}

To increase the recall rate of pseudo boxes, we repeat the above procedure multiple times with different threshold values $\tau$. The final results are merged with Non-Maximum Suppression (NMS) based on the Intersection over Union (IoU) between boxes. The boxes are assigned the ground-truth image-level labels as class labels.

\paragraph{Step 3: Detector Pre-training.}
The pseudo box generation procedure effectively transforms a classification dataset into a detection dataset to be readily used in a standard detection training algorithm. 
We initialize the backbone in this step with the classification model, and initialize the detector-specific components such as FPN, RPN, and detection heads randomly. Note that we intentionally simplify the pre-training step by treating the detector as a black box. This has two advantages: (1) The approach can be easily generalized to other detector architectures; (2) The approach can leverage existing knowledge about how to train those architectures well and requires minimal code change.

\paragraph{Step 4: Downstream Detector Fine-tuning.}
When fine-tuning the downstream detection tasks, the pre-trained detector weights are used to initialize a new model, except for the last layers which depend on the number of categories.
Our approach is able to initialize more network layers than the traditional classification pre-training.

\subsection{Discussion}

In Step 2, we adopt a straight-forward CAM-based WSOL approach for its simplicity. 
An alternative design choice is to obtain the localization information through WSOD \cite{bilen2016weakly,kantorov2016contextlocnet,tang2017multiple,wei2018ts2c,zeng2019wsod2,zhong2020boosting}. However, WSOD is computationally expensive for large-scale datasets, as it typically needs to extract hundreds or thousands of proposals (\eg, through Selective Search \cite{uijlings2013selective}) and learn a multi-instance classifier.
Handling cluttered scenes by WSOD is in general a hard problem that warrants further study.
In contrast, our approach takes advantage of simple scenes (in, \eg, ImageNet) and only needs to quickly scan each image in an inference mode without extra training, which can be easily scaled up to larger-scale datasets.

As the bounding boxes are not verified by a human judge, the pseudo annotation could be noisy, \eg, incomplete boxes, incorrect localization. 
However, the pseudo annotation is only used for pre-training, and the fine-tuning process can compensate for the noisy labels to a certain extent.
While a more sophisticated treatment might produce more accurate pseudo boxes, we find in the experiments that the pseudo boxes generated from our simple approach can yield substantial improvement in downstream detection tasks through detection-aware pre-training. 

\section{Experiment}\label{sec:exp}

\subsection{Settings}
\vspace{-10pt}

\paragraph{Pre-training Datasets.} 
We use ImageNet-1M and ImageNet-14M \cite{deng2009imagenet,russakovsky2015imagenet}
as the pre-training datasets. ImageNet-1M contains $1.28$ million images of $1K$ categories. ImageNet-14M is a larger variant of ImageNet which contains $14$ million images of $22K$ categories.

\paragraph{Detection Datasets.} For the detection fine-tuning tasks, we leverage the widely-used Pascal VOC \cite{everingham2010pascal}, Common Objects in Context (COCO) \cite{lin2014microsoft} datasets. The Pascal VOC dataset has different versions of each year's competition. Our first setting is based on the VOC 2007 version, where the training set is the trainval2007 ($5,011$ images) and the evaluation set is test2007 ($4,952$ images). The other setting, which we refer to as VOC 07+12, is to merge the trainval2007 and trainval2012 as the training set ($11,540$ images in total), and evaluate on the test2007 set. 
This is a widely-used protocol in the literature \cite{he2019rethinking}. The VOC dataset has $20$ object categories.
For the COCO dataset, we adopt the COCO 2017 train/val split where the train set contains $118K$ valid images and the val set has $5000$ images. The COCO dataset has $80$ object categories.
On top of the aforementioned settings, we also simulate the corresponding low-data settings by varying the number of randomly sampled per-class images (5, 10, 20, 50, 100 images per class), to compare the fine-tuning sample efficiency of different pre-training strategies.

\paragraph{Architecture.} 
Our approach is independent of the detector framework. Here, we use Faster RCNN \cite{ren2015faster} with Feature Pyramid Networks (FPN) \cite{lin2017feature} and ResNet-50 \cite{he2016deep} as the testbed. In ablation studies, we also include other variants, \eg, RetinaNet \cite{lin2017focal} and ResNet-101 backbone \cite{he2016deep}.

\paragraph{Hyper-parameters.} 
In the first stage of classifier pre-training, we use the torchvision\footnote{https://pytorch.org/docs/stable/torchvision/index.html} pre-trained model for ImageNet-1M experiments.  For ImageNet-14M, the classifier is trained with batch size as $8192$ for $50$ epochs on $64$ GPUs. The initial learning rate is $3.2$ and decayed with a cosine scheduler.

In the second stage of pseudo box generation, we average the CAMs obtained from two image scales, \ie, short side length as $288$ or $576$, and from the original and the left-right flipped images. 
On the normalized ($[0,1]$) CAMs, we use $4$ different thresholds, \ie, $\tau = 0.2$, $0.3$, $0.4$, $0.5$, to generate boxes of various sizes to improve the recall rate. 
In the end, the mined boxes are merged by NMS with IoU threshold $0.8$. The $\tau$ and the NMS IoU threshold are further studied in the supplementary material.
With ResNet-50, this stage takes less than $13$ min on ImageNet-1M and about $2.3$ hours on ImageNet-14M with $64$ GPUs.

In the third stage of detector pre-training, the model is trained with batch size $32$ on $16$ GPUs for $40,038$ iterations on ImageNet-1M or $443,658$ iterations on ImageNet-14M.
We enable multi-scale augmentation, \ie, the short edge is randomly drawn from $(96,160,320,640)$.
The smallest scale is as small as $96$ because ImageNet tends to contain large central objects, while we expect the pre-trained detector to be able to handle diverse object scales.
This stage takes roughly $1.8$ hours on ImageNet-1M or $17.6$ hours on ImageNet-14M with Faster RCNN FPN ResNet-50, which is only a small extra cost on top of classification pre-training. As a reference, $90$ epochs of ImageNet-1M ResNet-50 classifier training takes $7$ hours on $16$ GPUs.

In the final stage of fine-tuning, we perform experiments on Pascal VOC \cite{everingham2010pascal} and COCO \cite{lin2014microsoft}.
On COCO, the model is fine-tuned with $90K$ steps ($1$x) with batch size $16$.
The initial learning rate is $0.02$ and reduced by $0.1$ times at $60K$ and $80K$ steps. The image's short side is $800$.
On VOC 07 and VOC 07+12, the model is trained for $14$ epochs ($4.5K$ steps). 
The initial learning rate is $0.01$ and reduced to $0.001$ at the $10$th epoch. 
The input image's short side is $640$.

For the low data settings, training with the same number of iterations as the full data setting is sub-optimal. Early stop is needed. Following \cite{he2019rethinking}, we tune the number of iterations. As in \cite{he2017mask,massa2018mrcnn}, we use fixed BatchNorm and freeze the the first conv block of ResNet in all fine-tuning experiments. The weight decay coefficient is set to 1e-4 in ImageNet-1M experiments and 1e-5 in ImageNet-14M experiments. We do not use test time multi-scale augmentation.

\paragraph{Evaluation metrics.}
On COCO, we report the standard AP metrics \cite{lin2014microsoft}, \ie, AP\textsubscript{.5:.95}, the mean of average precisions (AP) evaluated at IoU thresholds $0.5$, $0.55$, $\cdots$, $0.95$. AP\textsubscript{.5} and AP\textsubscript{.75} are also reported for AP at IoU $0.5$ and $0.75$.
AP\textsubscript{\{s,m,l\}} are for small ($< 32^2$ pixels), medium, and large ($\ge 96^2$ pixels) objects, determined by the area of a bounding box.
For VOC, we report AP\textsubscript{.5} and the 11-point version AP\textsubscript{.5,07metric} defined by the VOC 2007 challenge \cite{everingham2010pascal}.

\subsection{Main results}

\begin{table}[tb]
    \centering
    \caption{COCO full-data detection results. CLS and DAP refer to the baseline classification and our pre-training strategies. The improvement of DAP over CLS is marked in $\Delta$ row. IN-1M and IN-14M correspond to using ImageNet-1M or ImageNet-14M as pre-training set. We report the AP\textsubscript{.5:.95}: the mean of average precisions, AP\textsubscript{.5}, AP\textsubscript{.75}: AP at IoU 0.5 and 0.75, AP\textsubscript{\{s,m,l\}}: AP for small, medium, large objects, calculated on COCO 2017 val.}
    \small
    \setlength{\tabcolsep}{4pt}
    \begin{tabular}{l|ccc|ccc}
        \toprule
        Pre-train  & AP\textsubscript{.5:.95} & AP\textsubscript{.5} & AP\textsubscript{.75} & AP\textsubscript{s} & AP\textsubscript{m} & AP\textsubscript{l} \\
        \midrule
        IN-1M CLS  & 36.73  & 58.04  & 39.72  & 20.57  & 39.56  & 48.51  \\
        IN-1M DAP  & 37.25  & 58.98  & 40.46  & 21.71  & 40.64  & 48.34  \\
        $\Delta$   & \gain{+0.52}  & \gain{+0.94}  & \gain{+0.74}  & \gain{+1.14}  & \gain{+1.08}  & \gain{+0.83}  \\
        \midrule
        IN-14M CLS & 38.87  & 61.87  & 42.41  & 23.79  & 42.15  & 49.89  \\
        IN-14M DAP & 39.57  & 63.05  & 43.02  & 24.03  & 42.96  & 51.15  \\
        $\Delta$   & \gain{+0.70}  & \gain{+1.18}  & \gain{+0.61}  & \gain{+0.24}  & \gain{+0.81}  & \gain{+1.26}  \\
        \bottomrule
    \end{tabular}
    \label{tab:r50_coco}
    \vspace{4pt}

    \small
    \caption{VOC 07 and 07+12 full-data detection results. CLS and DAP refer to the baseline classification and our pre-training strategies. IN-1M and IN-14M correspond to using ImageNet-1M or ImageNet-14M for pre-training. We report AP\textsubscript{.5} which is the area under the precision-recall curve at IoU threshold 0.5, and AP\textsubscript{.5, 07metric} which is the 11-point metric at IoU 0.5 defined in Pascal VOC 2007 challenge \cite{everingham2010pascal}, caculated on VOC 2007 test.}
    \small
    \setlength{\tabcolsep}{6.2pt}
    \begin{tabular}{llll}
        \toprule
        Train set & Pre-train & AP\textsubscript{.5} & AP\textsubscript{.5, 07metric} \\
        \midrule
        07 trainval & IN-1M CLS  & 77.36  & 75.00  \\
                    & IN-1M DAP  & 79.93 (\gain{+2.57})  & 77.57 (\gain{+2.57})  \\
                    & IN-14M CLS & 80.74  & 78.29  \\
                    & IN-14M DAP & 84.24 (\gain{+3.50})  & 81.54 (\gain{+3.25})  \\
        \midrule
        07+12 & IN-1M CLS  & 83.77  & 80.97  \\
        trainval       & IN-1M DAP  & 84.49 (\gain{+0.72}) & 82.00 (\gain{+1.03})  \\
                       & IN-14M CLS & 86.91  & 83.56  \\
                       & IN-14M DAP & 87.84 (\gain{+0.93}) & 84.53 (\gain{+0.97})  \\
        \bottomrule
    \end{tabular}
    \label{tab:r50_voc}
    \vspace{-10pt}
\end{table}

\begin{figure}[t]
    \centering 
    \includegraphics[width=\linewidth]{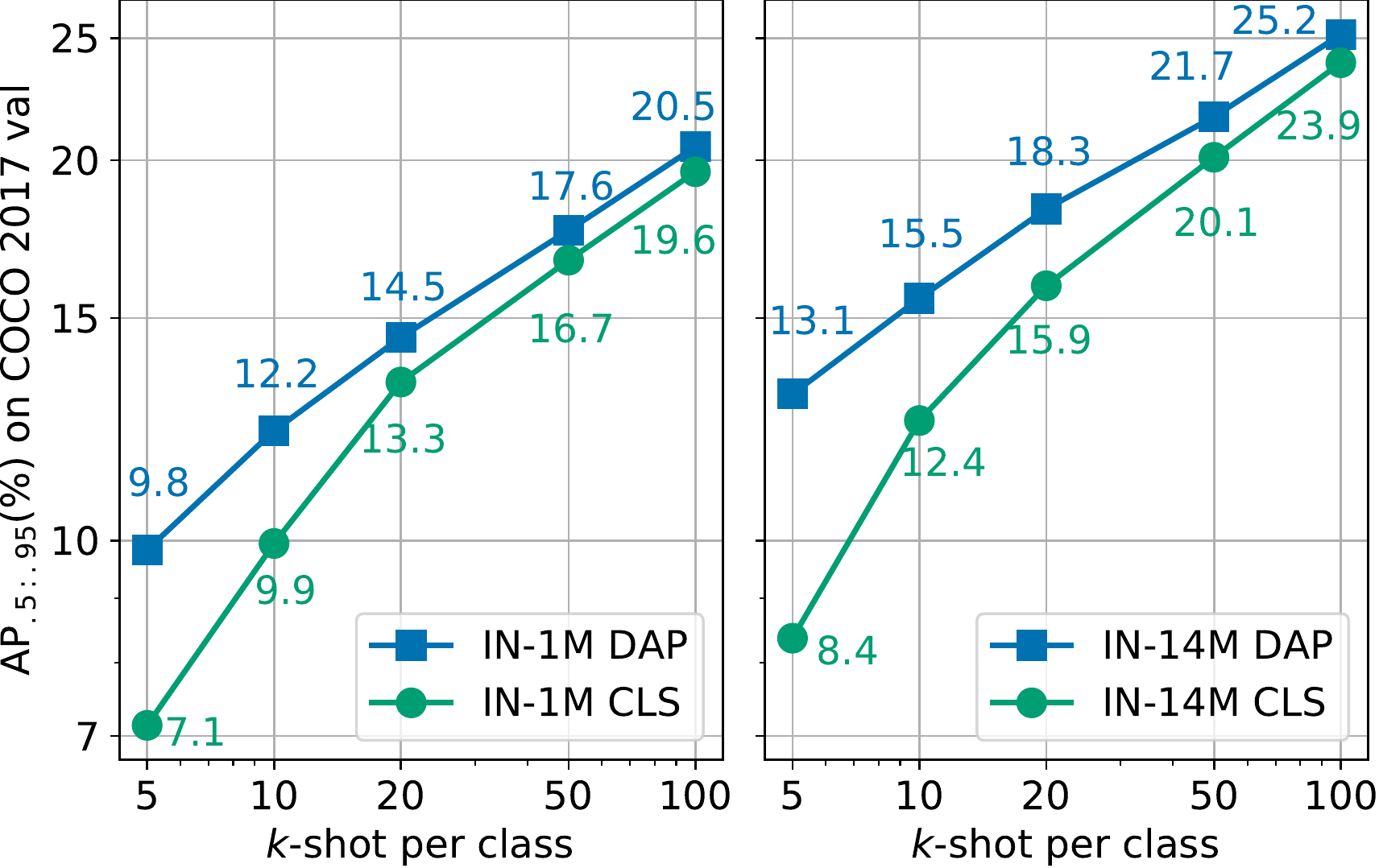}
    \caption{COCO $k$-shot low-data detection results. CLS and DAP refer to the baseline classification and our pre-training strategies. IN-1M (left) and IN-14M (right) correspond to using ImageNet-1M or ImageNet-14M as pre-training set. In the horizontal direction, we vary the number of images per class, and in the vertical direction, we report the AP\textsubscript{.5:.95} on COCO 2017 val. There are 80 classes in COCO, so 5-shot corresponds to 400 images in total.}
    \label{fig:coco}
\end{figure}

\begin{figure}[t]
    \centering
    \includegraphics[width=\linewidth]{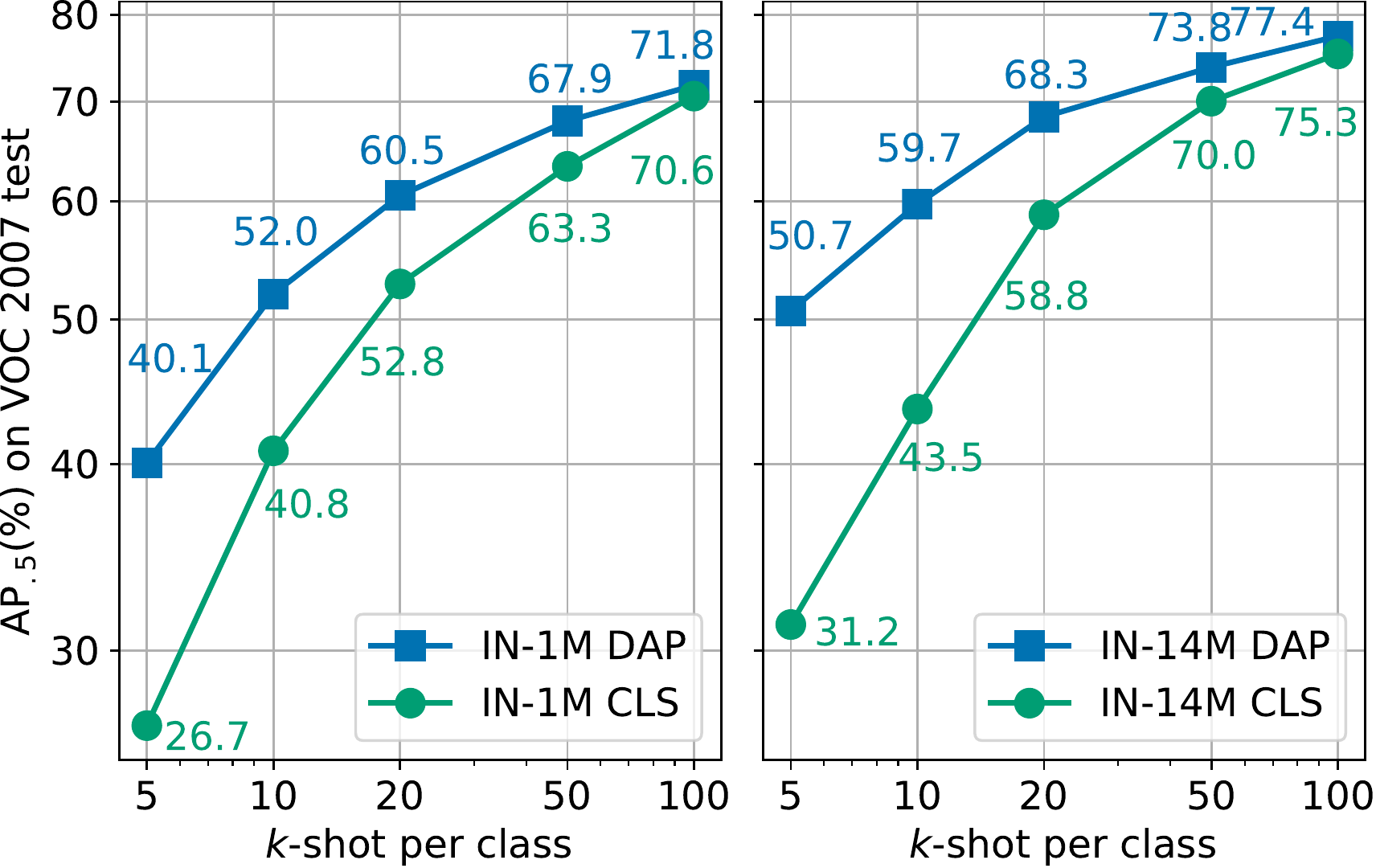}
    \caption{VOC $k$-shot low-data detection results. IN-1M (left) and IN-14M (right) refer to using ImageNet-1M or ImageNet-14M as pre-training set. In the horizontal direction, we vary the number of images per class, and in the vertical direction, we report the AP\textsubscript{.5} on VOC 2007 test. We sample the training images from the combined VOC 07+12 trainval set. There are 20 classes in VOC, so 5-shot corresponds to 100 images in total.}
    \label{fig:voc}
\end{figure}

We denote 1N-1M CLS and 1N-14 CLS as the short-hands for the traditional classification pre-training strategy on ImageNet-1M and ImageNet-14M, respectively. Similarly, our DAP strategy is denoted as IN-1M DAP and IN-14M DAP on the two ImageNet dataset variants. In DAP, 4.1M pseudo boxes are mined in ImageNet-1M and 47M boxes in ImageNet-14M.
The results are summarized in Tables~\ref{tab:r50_coco},~\ref{tab:r50_voc} for the full-data setting and Figures~\ref{fig:coco},~\ref{fig:voc} for the low-data setting, and observations are as follows.

\paragraph{DAP is more effective than classification pre-training in the full-data setting.}
The full-data setting results in Tables~\ref{tab:r50_coco} and \ref{tab:r50_voc} tell that DAP performs consistently better than classification pre-training (CLS) across all metrics. The gain is especially significant for the VOC dataset, reaching a $\ge2.5$ AP\textsubscript{.5} increase with 07 trainval and a roughly $+1$ AP\textsubscript{.5} increase with 07+12 trainval. 
And the gain on COCO AP\textsubscript{.5:.95} is $+0.52$ with ImageNet-1M and $+0.7$ with ImageNet-14M.
The results suggest that DAP makes better use of the ImageNet dataset to pre-train the network than CLS pre-training.

\paragraph{DAP benefits more from larger pre-training dataset.}
Comparing ImageNet-1M and ImageNet-14M, our DAP scales up better to ImageNet-14M. Improvement on ImageNet-14M is larger than on ImageNet-1M: $+0.7$ vs $+0.52$ on $\Delta$AP\textsubscript{.5:.95} with COCO in Table~\ref{tab:r50_coco}, $+3.5$ vs $+2.16$ with VOC 07 and $+0.93$ vs $+0.72$ with VOC 07+12 on $\Delta$AP\textsubscript{.5} in Table~\ref{tab:r50_voc}. The training process and pseudo box generation hyper-parameters are shared between the 1M and 14M results. The only difference is the size of the pre-training datasets. Therefore, this observation suggests that DAP benefits more from the larger ImageNet dataset.

\paragraph{DAP improves low-data performance.}
The low-data setting is of great practical value to reduce the annotation cost.
In Figure~\ref{fig:coco}, \ref{fig:voc}, we mimic this low-data regime by downsampling the COCO and VOC datasets.
Compared with CLS, we observe that fine-tuning from DAP benefit much more in the low-data setting than the full-data setting. For example, in the 5-shot case in Figure~\ref{fig:coco}, IN-1M DAP outperforms IN-1M CLS pre-training by $2.6$ AP\textsubscript{.5:.95} (left), and IN-14M DAP surpasses IN-14M CLS by a significant $4.7$ AP\textsubscript{.5:.95} (right). Similarly, in Figure~\ref{fig:voc}, the VOC $\Delta$AP\textsubscript{.5} is as much as $+13.4$ (IN-1M) and $+19.5$ (IN-14M) in the 5-shot case, compared to $+0.72$ and $+0.93$ in the full-data setting.

\subsection{Analysis}
\vspace{-8pt}

\begin{figure}
    \centering
    \includegraphics[width=\linewidth]{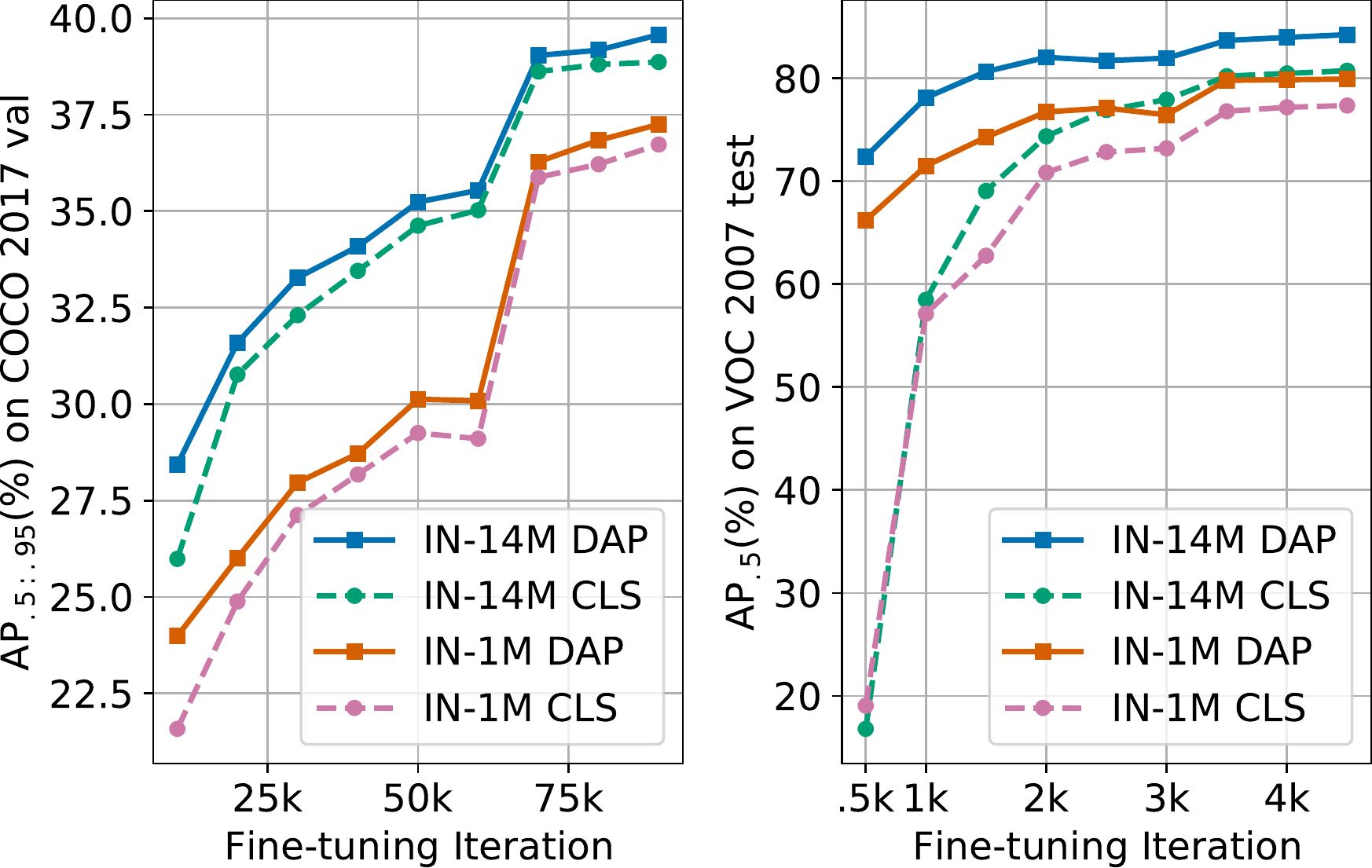}
    \caption{Learning curves of Faster RCNN FPN ResNet-50 during COCO and VOC 07 training, with different pre-trained models as initialization. DAP is able to provide a ``head-start'' for fine-tuning (i.e., faster convergence), and is almost always leading the CLS pre-trained counterparts. The abrupt increase in COCO AP is caused by learning rate reduction at $60$k-step.}
    \label{fig:converge}
    \vspace{-8pt}
\end{figure}

\paragraph{Faster convergence with DAP than classification pre-training.}
As our DAP approach provides greater accuracy improvement, we study the convergence behavior by plotting the learning curves of COCO and VOC training with different pre-trained ResNet-50 models in Figure~\ref{fig:converge}.
From the figure, we notice that DAP can give a significant initial accuracy boost compared to CLS. For example, in the right part of Figure~\ref{fig:converge}, $500$ fine-tuning iterations from DAP already achieve $\ge 65$ AP\textsubscript{.5}, while the corresponding CLS numbers are lower than $20$. 
This demonstrates that a better pre-trained model can provide faster convergence speed, which is consistent with \cite{mahajan2018exploring, kolesnikov2019big,he2019rethinking,shinya2019understanding}.

\begin{figure}
    \centering
    \includegraphics[width=\linewidth]{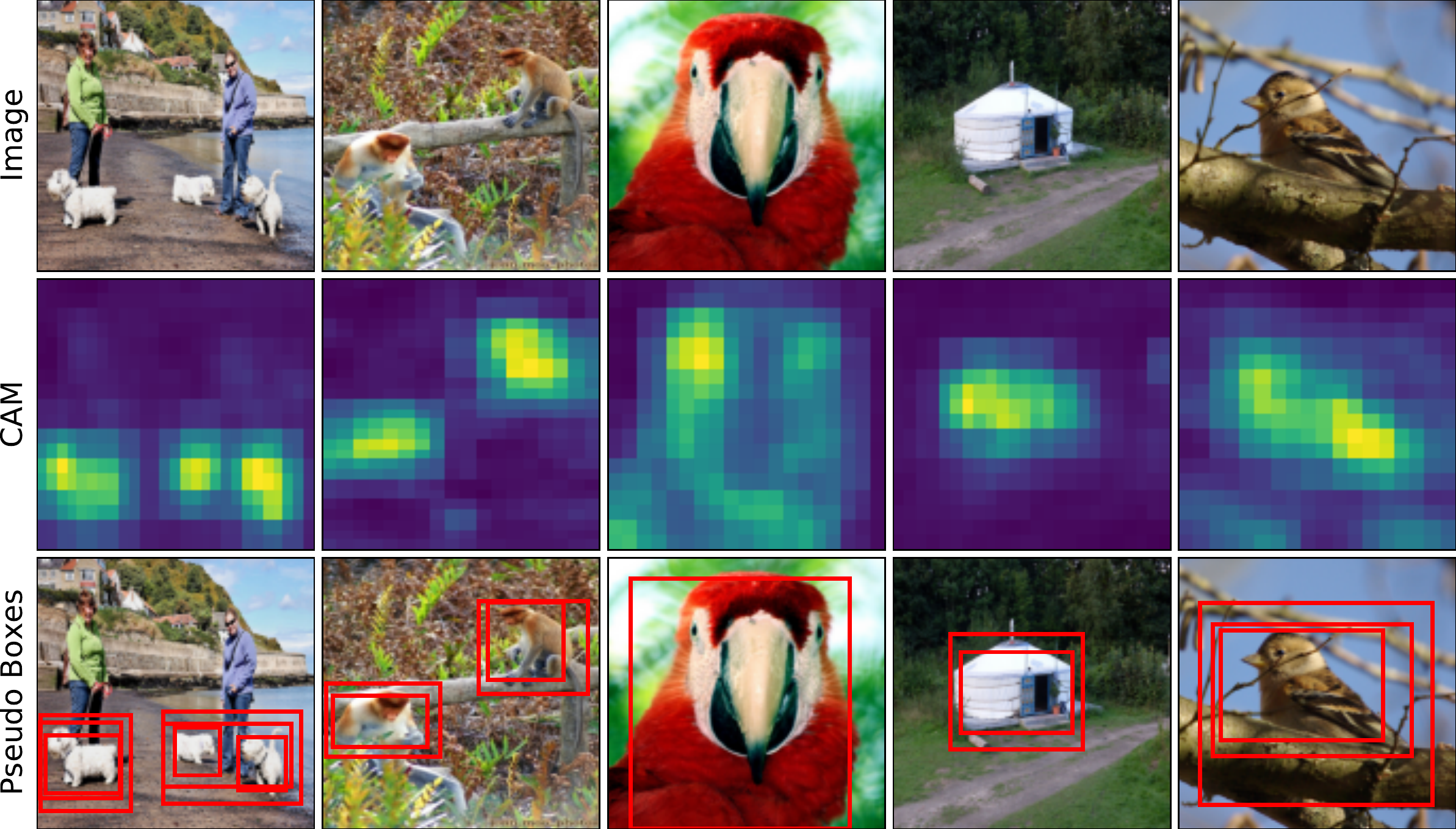}
    \caption{Examples of ImageNet images, CAMs \& pseudo boxes.}
    \label{fig:pseudo_box}
\end{figure}

\paragraph{Visualization of CAM pseudo boxes.}
Figure~\ref{fig:pseudo_box} visualizes the CAMs and the mined pseudo boxes in ImageNet-1M images. In all examples, our pseudo box generation procedure can successfully find the rough locations of the objects. The per-component multi-threshold approach is able to recover multiple objects in the first two columns. We notice that the pseudo boxes are noisy, i.e., containing inaccurate boxes such as the loose one around the bird in the last column. Despite the noise, pre-training can still benefit from leveraging a large amount of data. The network can pick up useful learning signals from the noisy pseudo labels.

\paragraph{What is learned in pre-training?}
To study whether the better performance of DAP is only due to being able to pre-train the additional components of detectors, we conduct an ablation study by freezing the whole ResNet-50 backbone and only pre-training the FPN, RPN and ROI heads in Faster RCNN. The result is in Table~\ref{tab:freeze}. The COCO AP\textsubscript{.5:.95} in this setting is $36.89$, and AP\textsubscript{.5} is $78.86$ when trained on VOC 07 and $83.91$ trained on VOC 07+12. The downstream task performance is better than CLS pre-training but worse than full DAP, suggesting that DAP not only pre-trains the new layers, but also adapts the feature representations of the entire network more towards detection.

\begin{table}[t]
    \centering
    \small
    \caption{Comparison to the variant that freezes ResNet-50 backbone and only pre-trains the additional layers in Faster RCNN.}
    \setlength{\tabcolsep}{3.92pt}
    \begin{tabular}{llll}
        \toprule
        Pre-train & \scriptsize{COCO AP\textsubscript{.5:.95}} &  \scriptsize{VOC07 AP\textsubscript{.5}}  & \scriptsize{VOC07+12 AP\textsubscript{.5}}  \\
        \midrule
        IN-1M CLS        & 36.73  & 77.36  & 83.77  \\
        IN-1M DAP        & 37.25  & 79.93  & 84.49  \\
        IN-1M DAP (Freeze)
                         & 36.89  & 78.86  & 83.91  \\
        \bottomrule
    \end{tabular}
    \label{tab:freeze}
    \vspace{-5pt}
\end{table}

\begin{table}[tb]
    \centering
    \caption{ResNet-101 Faster RCNN FPN results on COCO.}
    \small
    \setlength{\tabcolsep}{3.92pt}
    \begin{tabular}{l|ccc|ccc}
        \toprule
        Pre-train  & AP\textsubscript{.5:.95} & AP\textsubscript{.5} & AP\textsubscript{.75} & AP\textsubscript{s} & AP\textsubscript{m} & AP\textsubscript{l} \\
        \midrule
        IN-1M CLS  & 39.11  & 61.06  & 42.59  & 22.98  & 42.35  & 50.50  \\
        IN-1M DAP  & 39.28  & 61.32  & 42.81  & 23.45  & 42.70  & 51.47  \\
        $\Delta$   & \gain{+0.17} & \gain{+0.26} & \gain{+0.22} & \gain{+0.47} & \gain{+0.35} & \gain{+0.97}  \\
        \midrule
        IN-14M CLS & 43.18  & 66.76  & 47.31  & 26.29  & 47.21  & 55.55  \\
        IN-14M DAP & 43.92  & 67.16  & 48.39  & 27.41  & 48.18  & 56.37  \\
        $\Delta$   & \gain{+0.74}  & \gain{+0.40}  & \gain{+1.08}  & \gain{+1.12}  & \gain{+0.97}  & \gain{+0.82}  \\
        \bottomrule
    \end{tabular}
    \label{tab:r101_coco}
    \vspace{4pt}
    
    \caption{ResNet-101 Faster RCNN FPN results on VOC.}
    \small
    \setlength{\tabcolsep}{6pt}
    \begin{tabular}{llll}
        \toprule
        Train set & Pre-train & AP\textsubscript{.5} & AP\textsubscript{.5, 07metric} \\
        \midrule
        07 trainval & IN-1M CLS  & 78.02  & 75.73  \\
                    & IN-1M DAP  & 80.95 (\gain{+2.93})  & 78.03 (\gain{+2.30})  \\
                    & IN-14M CLS & 84.58  & 81.88  \\
                    & IN-14M DAP & 86.63 (\gain{+2.05})  & 83.71 (\gain{+1.83})  \\
        \midrule
        07+12          & IN-1M CLS  & 84.91  & 81.81  \\
        trainval       & IN-1M DAP  & 85.49 (\gain{+0.58})  & 82.43 (\gain{+0.62})  \\
                       & IN-14M CLS & 89.41  & 85.92  \\
                       & IN-14M DAP & 90.69 (\gain{+1.28})  & 86.55 (\gain{+0.63}) \\
        \bottomrule
    \end{tabular}
    \label{tab:r101_voc}
\end{table}

\paragraph{Varying network backbone}
We change the backbone network from ResNet-50 to the larger ResNet-101. The results are in Table~\ref{tab:r101_coco}, \ref{tab:r101_voc} for both COCO and VOC. 
ResNet-101 delivers higher absolute accuracy than Resnet-50, but DAP again performs consistently better than CLS pre-training.

\begin{table}[tb]
    \centering
    \caption{RetinaNet (ResNet-50 FPN) results on COCO 2017 val.}
    \small
    \setlength{\tabcolsep}{4.2pt}
    \begin{tabular}{l|ccc|ccc}
        \toprule
        Pre-train  & AP\textsubscript{.5:.95} & AP\textsubscript{.5} & AP\textsubscript{.75} & AP\textsubscript{s} & AP\textsubscript{m} & AP\textsubscript{l} \\
        \midrule
        IN-1M CLS  & 36.22  & 55.11  & 38.56  & 19.72  & 39.56  & 48.74  \\
        IN-1M DAP  & 36.96  & 55.99  & 39.43  & 20.41  & 40.20  & 50.61  \\
        $\Delta$ & \gain{+0.74} & \gain{+0.88} & \gain{+0.87} & \gain{+0.69} & \gain{+0.64} & \gain{+1.87}  \\
        \bottomrule
    \end{tabular}
    \label{tab:retinanet_coco}
    \vspace{4pt}
    \caption{RetinaNet (ResNet-50 FPN) results on VOC 2007 test.}
    \small
    \setlength{\tabcolsep}{6.5pt}
    \begin{tabular}{llll}
        \toprule
        Train set & Pre-train & AP\textsubscript{.5} & AP\textsubscript{.5, 07metric} \\
        \midrule
        07 trainval & IN-1M CLS  &  75.12  & 72.91  \\
                    & IN-1M DAP  &  77.95 (\gain{+2.83})  & 75.80 (\gain{+2.89}) \\
        \midrule
        07+12       & IN-1M CLS  & 81.48  & 78.69  \\
        trainval    & IN-1M DAP  & 84.18 (\gain{+2.70}) & 81.15 (\gain{+2.46}) \\
        \bottomrule
    \end{tabular}
    \label{tab:retinanet_voc}
    \vspace{-8pt}
\end{table}

\paragraph{Varying detector architecture.}
Our DAP requires no knowledge of the internal mechanism of a detector.
We show that our DAP approach generalizes to the RetinaNet detector architecture \cite{lin2017focal}. RetinaNet is a one-stage detector as opposed to the two-stage detector of Faster RCNN. We pre-train a RetinaNet (ResNet-50) detector on IN-1M for $80,072$ steps with batch size $32$ and learning rate $0.005$. We can see in Table~\ref{tab:retinanet_coco}, \ref{tab:retinanet_voc} that the same pipeline works well with RetinaNet and DAP consistently outperforms CLS pre-training on both COCO and VOC.

\paragraph{Accuracy on ImageNet.}
To study the effect of DAP on the original ImageNet classification task, we evaluate the IN-1M DAP pre-trained Faster RCNN (ResNet-50) as a classifier. The class score is taken as the sum of the confidence scores of all detected objects of this class. The DAP pre-trained detector achieves $62.73\%$ Top-1 accuracy and $85.99\%$ Top-5 accuracy. These numbers are lower than those of the bare ResNet-50 backbone ($76.15\%$ Top-1 and $92.87\%$ Top-5). However, as our experiments have shown, the DAP pre-trained network is better at detection fine-tuning, suggesting that the drop in whole-image classification accuracy is likely traded for better localization and regional classification ability.

\paragraph{Reference with IN-200 DET.}
ImageNet challenge provides a detection subset of $200$ classes, which is referred  as IN-200 DET \cite{deng2009imagenet,russakovsky2015imagenet}. We present a reference result on using this dataset in fully-supervised detection pre-training in Table~\ref{tab:in200}. We train on IN-200 DET for 5 epochs with scale augmentation $(96,160,320,640)$ and transfer the detector weights. The VOC07 AP\textsubscript{.5} is improved to $80.53$ (vs CLS pre-training $77.36$) when trained on VOC 07 and $84.01$ (vs CLS $83.77$) trained on VOC 07+12.
Our DAP achieves even higher accuracy except for the slight drop with IN-1M VOC07. 
This suggests that DAP may benefit from the substantially more pseudo boxes mined from a larger-scale dataset than the human-labeled boxes in IN-200 DET.

\begin{table}[t]
    \centering
    \small
    \caption{ImageNet-200 DET reference results trained on VOC 07 trainval and 07+12 trainval, evaluated on VOC 2007 test.}
    \setlength{\tabcolsep}{2.4pt}
    \begin{tabular}{lccccc}
        \toprule
        Pre-train   &  \#Image & \#Class & \#Box & \scriptsize{VOC07 AP\textsubscript{.5}} & \scriptsize{VOC07+12 AP\textsubscript{.5}}  \\
        \midrule
        IN-1M CLS   & 1.28M & 1K  & 0     & 77.36  & 83.77  \\
        IN-1M DAP   & 1.28M & 1K  & 4.11M & 79.93  & 84.49  \\
        IN-14M CLS  & 14.2M & 22K & 0     & 80.74  & 86.91  \\
        IN-14M DAP  & 14.2M & 22K & 47.3M & 84.24  & 87.84  \\
        IN-200 DET  & 333K  & 200 & 479K  & 80.53  & 84.01 \\
        \bottomrule
    \end{tabular}
    \label{tab:in200}
\end{table}

\begin{table}[tb]
    \centering
    \caption{Faster RCNN FPN ResNet-50 results of training on VOC 07 trainval + self-training on 2012 trainval, evaluated on 07 test.}
    \small
    \setlength{\tabcolsep}{6.6pt}
    \begin{tabular}{lcll}
        \toprule
        Pre-train & +Self-train & AP\textsubscript{.5} & AP\textsubscript{.5, 07metric} \\
        \midrule
        IN-1M CLS  &             & 77.36  & 75.00  \\
        IN-1M DAP  &             & 79.52 (\gain{+2.16}) & 76.82 (\gain{+1.82}) \\
        \midrule
        IN-1M CLS  & \checkmark  & 77.98  & 75.71  \\
        IN-1M DAP  & \checkmark  & 80.50 (\gain{+2.52})  & 78.02 (\gain{+2.31})  \\
        \bottomrule
    \end{tabular}
    \label{tab:self-train}
    \vspace{-8pt}
\end{table}

\paragraph{Complementary to self-training.}
Self-training is another direction of sample-efficient learning which re-trains the model with unlabeled data and pseudo labels \cite{radosavovic2018data,zoph2020rethinking,xu2020self}.
We believe self-training and pre-training are complementary. Intuitively, better pre-training may give the model a head-start in learning downstream tasks, therefore produce better pseudo labels for the subsequent self-training.
We consider a VOC semi-supervised setting and verify that
DAP can still improve the accuracy under self-training settings.
The detector is trained on VOC 2007 trainval initialized with our DAP or with the conventional CLS.
Then, We keep as pseudo labels the confident predictions, \ie, score $\ge0.6$ or is the max in the image on the VOC 2012 trainval (ground-truth labels are removed to have a semi-supervised setting) with test-time flip augmentation, and finally tune the detector for 2 more epochs on all data. 
The result is shown in Table~\ref{tab:self-train}. 
Self-training improves AP across all pre-training strategies. 
Notably, in this particular setting, DAP leads to even larger gain than CLS.
That is, without self-training, our improvement is $+2.16$ AP\textsubscript{.5} and with self-training, the gain is $+2.52$.

\paragraph{Longer Training Schedule.}
In Table~\ref{tab:long}, we ran 3 settings (Faster RCNN ResNet-50) for $3$\texttt{x} longer to study whether the gain of DAP persists with longer fune-tuning time. We observe the difference between DAP and CLS on COCO full gets smaller. However, DAP still brings noticeable gains in the 5-shot settings that are prone to overfitting.

\begin{table}[tb]
\centering
\caption{Comparing CLS and DAP with longer fine-tuning time.}
\label{tab:long}
\footnotesize
\setlength{\tabcolsep}{2pt}
\begin{tabular}{@{}l|cc|cc|cc}
    \toprule
       & {\scriptsize voc07 $5$shot $1$\texttt{x}} & {\scriptsize $3$\texttt{x}}
       & {\scriptsize coco $5$shot $1$\texttt{x}} & {\scriptsize $3$\texttt{x}}
       & {\scriptsize coco full $1$\texttt{x}} & {\scriptsize $3$\texttt{x}}  \\
    \hline
    {\footnotesize IN-1M CLS}  & 26.7  & 27.3   & 7.13  & 7.03   & 36.73  & 37.24  \\
    {\footnotesize IN-1M DAP}  & 40.1  & 40.9   & 9.88  & 9.20   & 37.25  & 37.45  \\
    \bottomrule
\end{tabular}
\vspace{-8pt}
\end{table}

\section{Discussion and Conclusion}
\vspace{-8pt}

\paragraph{Implication for future work.} This work may open up many future directions. We adopt a straightforward WSOL method in this paper. A more sophisticated WSOL or WSOD method \cite{bilen2016weakly,kantorov2016contextlocnet,tang2017multiple,wei2018ts2c,zeng2019wsod2} could potentially produce higher-quality pseudo boxes, which may improve pre-training in return. For example, it may require handling the missing label problem in ImageNet. However, we want to emphasize that the simplicity of our pseudo box generation method is also a blessing by being scalable to millions of images such as ImageNet-14M.
Another sensible next step is to leverage mixed-labeled data in pre-training, e.g., using semi-supervised WSOD as the pre-training procedure \cite{zhong2020boosting}.
DAP might also benefit from moving to a more diverse multi-labeled dataset.
This would make object localization more challenging, but the network may benefit from seeing more complex contexts. Finally, broadening the approach to mask detection (or instance segmentation) \cite{he2017mask} aware pre-training is worth exploring.

\paragraph{Conclusions.}
In this paper, we have proposed a Detection-Aware Pre-training (DAP) strategy under weak supervision.
Specifically designed for object detection downstream tasks, DAP makes better use of a classification-style dataset by transforming it into a detection dataset through a pseudo box generation procedure.
The generation is based on a simple yet effective approach built on CAM.
DAP reduces the discrepancies in the objective function, the localization information, and the network structure between the pre-training and the fine-tuning tasks. As a consequence, DAP leads to faster and better fine-tuning than classification pre-training. 
Besides, DAP leads to much higher accuracy in low-data settings and is complementary to advances in self-training \cite{zoph2020rethinking}.


{\small
\bibliographystyle{ieee_fullname}
\bibliography{ref}
}

\appendix
\clearpage
\setcounter{figure}{0}
\renewcommand{\thefigure}{A\arabic{figure}}
\setcounter{table}{0}
\renewcommand{\thetable}{A\arabic{table}}

\section{Hyper-parameters and more visualization}

During the pseudo box generation, we adopt the multi-threshold ($0.2$, $0.3$, $0.4$, $0.5$) CAM thresholding strategy and choose the post-processing NMS IoU threshold as $0.8$. 
Here we study how these hyper-parameter values impact the pseudo boxes and the downstream detection performance under the IN-1M DAP Faster RCNN ResNet-50 setting.

\begin{table}[b]
    \small
    \centering
    \caption{Effect of the CAM Threshold $\tau$ on pseudo box generation and downstream task performance (NMS IoU=$0.8$) under the IN-1M DAP Faster RCNN ResNet-50 setting. The upper three rows show the average number of pseudo boxes per image, the average pseudo box width and height on ImageNet-1M. The bottom three rows show the downstream detection accuracy. ``Multi'' refers to the multiple-threshold strategy reported in the paper, which merges the box results from the 4 different thresholds. Other columns represent using a single threshold. We notice that a larger $\tau$ yields smaller boxes and ``Multi'' leads to the highest COCO AP.}
    \setlength{\tabcolsep}{5.7pt}
    \begin{tabular}{l|cccc|l}
        \toprule
        CAM Threshold $\tau$  & $0.2$  & $0.3$  & $0.4$  & $0.5$  & Multi  \\
        \midrule
        Avg boxes / image & 1.019 & 1.046 & 1.096 & 1.161  & 3.211
        \\
        Avg box width  & 339.9 & 296.5 & 250.4 & 199.6 & 253.4 \\
        Avg box height & 280.3 & 242.1 & 202.0 & 160.7 & 208.0
        \\
        \midrule
        COCO AP\textsubscript{.5:.95} & 36.89 & 36.82 & 36.90 & 36.97 & \textbf{37.25}
        \\
        VOC07 AP\textsubscript{.5}    & 79.47 & 79.57 & 79.13 & 79.06 & \textbf{79.93}
        \\
        VOC07+12 AP\textsubscript{.5} & 84.35 & 84.41 & \textbf{84.76} & 84.61 & 84.49
        \\
        \bottomrule
    \end{tabular}
    \label{tab:cam_thresh}
\end{table}

\begin{figure*}[hp]
    \centering
    \includegraphics[width=\linewidth]{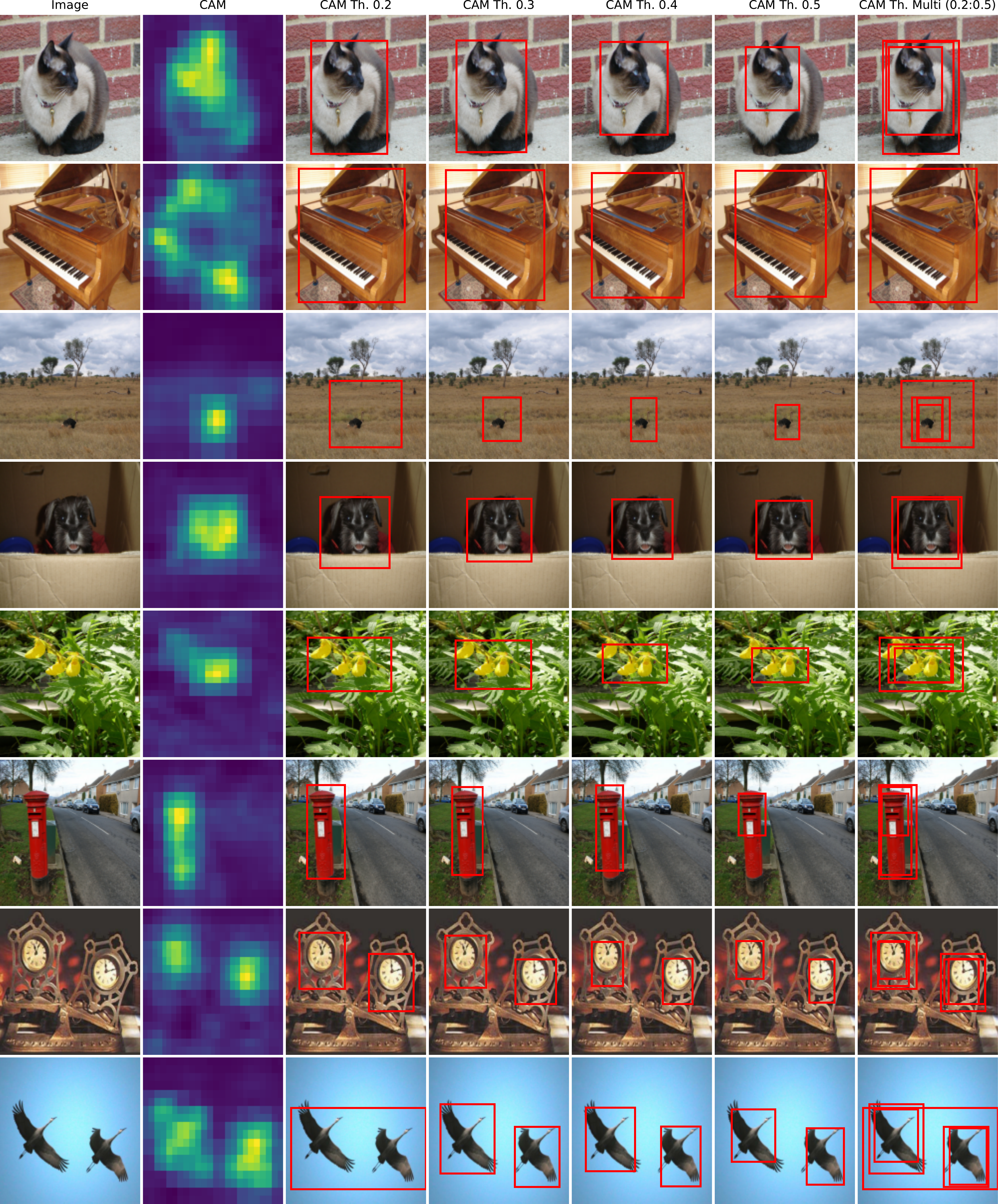}
    \caption{Visualization of pseudo boxes generated on ImageNet-1M with different CAM thresholds. In Row 1 and 6, lower threshold leads to more accurate boxes, while in Row 3, higher threshold can produce a tighter box. In the last row, higher threshold is required to discern the two birds. The final strategy in the main paper is ``Multi'', which combines boxes from different thresholds to improve recall.}
    \label{fig:vis_th}
\end{figure*}

\paragraph{CAM threshold $\tau$ for finding salient regions.}
Threshold $\tau$ separates out the salient regions on the class activation maps as in Eq.~(2) of the main paper. A higher $\tau$ may focus on the most activated regions, producing smaller bounding box instances on a single image.
On the other hand, a lower $\tau$ may keep more parts of the objects, producing larger boxes and more complete objects. This is shown more clearly in Table~\ref{tab:cam_thresh} and Figure~\ref{fig:vis_th}. As $\tau$ increases, the average number of mined objects per image increases, and the average size of boxes shrinks. The Multi (short for multiple-threshold) strategy introduced in the main paper is able to generate more pseudo boxes.

As for the downstream task accuracy, none of the single-threshold COCO results could match that of the multi-threshold strategy ($\ge 0.25$ AP drop), while for VOC07+12, the results of single threshold $0.4$ and $0.5$ are slightly better than Multi. COCO contains objects of diverse scales, while VOC contains mostly large objects only. This suggests that the multi-threshold strategy to find multiple pseudo boxes of different sizes is necessary to boost performance on COCO.

\begin{figure*}[hp]
    \centering
    \includegraphics[width=\linewidth]{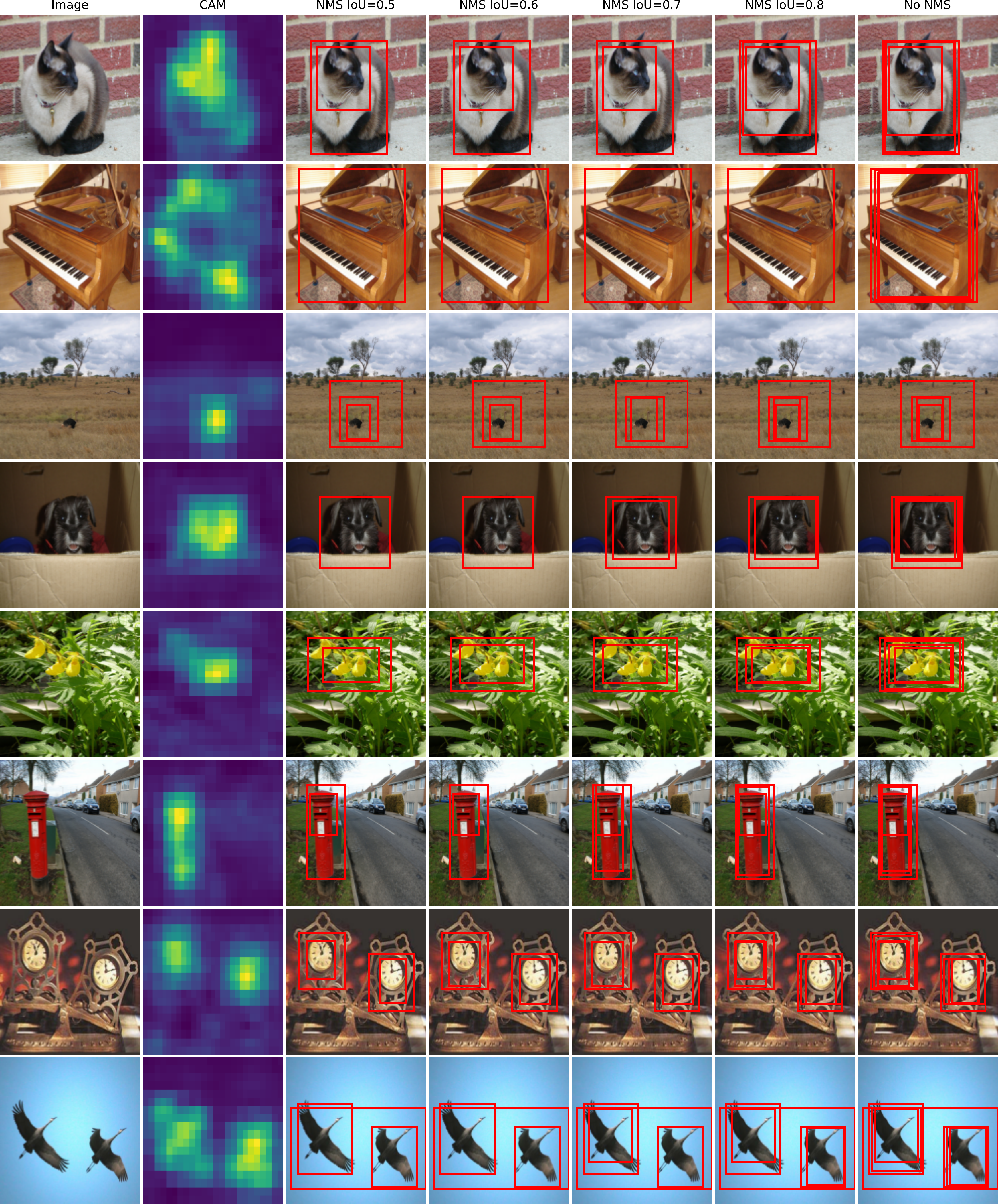}
    \caption{Visualization of pseudo boxes generated on ImageNet-1M with different NMS post-processing IoU thresholds. A smaller IoU threshold leads to fewer boxes. In the main paper, we set IoU as $0.8$ to keep more boxes since it yields good overall downstream detection performance on COCO and VOC.}
    \label{fig:vis_nms}
\end{figure*}

\paragraph{NMS IoU threshold for pseudo box post-processing.}
The pseudo boxes generated from different connected components with multiple CAM thresholds are merged with non-maximum suppression (NMS) post-processing. The NMS has an IoU threshold hyper-parameter. We vary the IoU threshold from $0.5$ to $1.0$ (no NMS) to study its effect on the pseudo box statistics and the downstream task performance, as shown in Table~\ref{tab:nms_thresh}. The boxes from different thresholds are visualized in Figure~\ref{fig:vis_nms}. We use the multi-threshold CAM strategy.
A higher IoU threshold keeps more pseudo boxes, while a lower threshold eliminates more boxes. Different IoUs do not have a significant impact on the size of the pseudo boxes.

In terms of the downstream task accuracy, an IoU threshold of $0.8$ or $0.9$ achieves the best AP on COCO. On VOC07, IoU $0.8$ achieves the best result, and on VOC07+12, IoU ranging from $0.6$ to $1.0$ gets similar AP\textsubscript{.5}. Since IoU $0.8$ delivers overall good performance while keeping only $3.2$ pseudo boxes per image, we choose this value in the main experiments.

\begin{table}[t]
    \small
    \centering
    \caption{Effect of NMS IoU threshold on pseudo box generation and downstream task performance under the IN-1M DAP Faster RCNN ResNet-50 setting. We vary the IoU threshold from $0.5$ to $1.0$. The IoU $0.8$ column is the one reported in the main paper. IoU $1.0$ refers to not doing the NMS post-processing.}
    \setlength{\tabcolsep}{3.4pt}
    \begin{tabular}{l|cccccc}
        \toprule
        NMS IoU   & $0.5$  & $0.6$  & $0.7$  & $0.8$  & $0.9$  & $1.0$  \\
        \midrule
        Avg boxes / image & 2.047 & 2.308 & 2.661 & 3.211 & 3.966 & 4.321
        \\
        Avg box width  & 258.9 & 256.2 & 254.2 & 253.4 & 260.9 & 269.6 \\
        Avg box height & 215.6 & 212.3 & 209.7 & 208.0 & 212.3 & 219.3
        \\
        \midrule
        COCO AP\textsubscript{.5:.95} & 37.12 & 36.98 & 37.13 & \textbf{37.25} & \textbf{37.26} & 37.02
        \\
        VOC07 AP\textsubscript{.5}    & 79.50 & 79.41 & 79.49 & \textbf{79.93} & 79.39 & 79.51
        \\
        VOC07+12 AP\textsubscript{.5} & 84.09 & \textbf{84.54} & 84.30 & 84.49 & 84.32 & 84.36
        \\
        \bottomrule
    \end{tabular}
    \label{tab:nms_thresh}
\end{table}

\end{document}